\documentclass[wcp]{jmlr}

\usepackage{longtable}

\usepackage{booktabs}

\usepackage{tikz}
\usetikzlibrary{arrows.meta,positioning,shapes.geometric,calc,fit,backgrounds,decorations.pathmorphing}
\usepackage{xcolor}
\usepackage{lineno}

\makeatletter
\let\Ginclude@graphics\@org@Ginclude@graphics
\makeatother

\jmlrvolume{353,}
\jmlryear{2026}
\jmlrworkshop{ACML 2026}

\title[ActionRAG]{ActionRAG: Evidence-Refining Graph Diffusion for Long-Term Memory in LLM Agents}

\author{\Name{Navyansh Mahla} \Email{navyansh@actioneer.com}\\
  \Name{Sashank Vandrangi} \Email{sashank@actioneer.com}\\
  \addr Actioneer, Mountain View, California, USA}

\editors{Andy Song, Bo Han and Sarah Erfani}

\begin{document}

\makeatletter
\let \@jmlrpages \@empty
\makeatother

\maketitle

\begin{abstract}
LLM agents need long-term memory that can recover precise and mutually
consistent evidence from growing interaction histories, particularly when
answers depend on relations among facts rather than isolated semantic
matches. Current memory pipelines remain vulnerable at two stages:
indexing errors can fragment repeated entities or merge distinct ones,
and online retrieval often treats initial similarity scores as final even
when the surrounding graph provides evidence for revising them. We
introduce \textbf{ActionRAG}, a graph memory architecture that couples
cautious indexing with query-time graph retroevaluation. Its indexing
pipeline reconciles type labels and applies tiered entity resolution.
Given a query, dual-channel entity and triple retrieval forms an initial
evidence set and induces a bounded local graph. Query-conditioned
receiver-normalized diffusion refines triple scores using local path
structure. Seed protection, hub-penalized flux, and triple
materialization preserve direct evidence while recovering missed
connections. We evaluate ActionRAG on
MemoryAgentBench (MAB), which tests accurate retrieval, conflict resolution,
long-range understanding, and test-time learning. Across its eight
deterministic-scoring columns, ActionRAG achieves
$55.1$ average accuracy, improving over the strongest retrieval-based
memory baseline by $5.4$ points. Under MAB's query-time cost accounting,
mean synthesis cost stays below \$0.001 per query on the four reported
cost columns, and the sum of separately measured retrieval-side $p_{99}$
components is
$440.7$\,ms on our consumer Apple Silicon setup.
\end{abstract}

\section{Introduction}

Large language model (LLM) agents accumulate information across many
turns of interaction and must recall, reconcile, and update facts as
those interactions unfold. A useful memory for such agents is not a
single skill but a bundle of them: looking up a specific fact buried
deep in past dialogue, reasoning across several facts to answer a
multi-hop question, recognizing when a new statement supersedes an
older one, and absorbing fresh information at test time. MemoryAgentBench
(MAB) \citep{hu2025memoryagentbench} operationalizes exactly this view,
evaluating memory in LLM agents via incremental multi-turn interactions
along four competencies: accurate retrieval, test-time learning,
long-range understanding, and conflict resolution.

The dominant memory architecture today is dense retrieval-augmented
generation (RAG) \citep{lewis2020rag}: chunk the interaction history,
embed each chunk, and look up the top-$k$ matches per query. This flat
index works for single-hop fact recall but loses signal on multi-hop
reasoning (each chunk is scored in isolation) and on conflict resolution
(a superseded fact and the fact that overrides it share most of their
semantic content and become indistinguishable to a chunk embedder). The
result is a measurable accuracy gap on exactly the MemoryAgentBench
slices that real agents need to handle.

Graph-structured memory addresses this by representing the agent's
history as a typed entity--relation graph. Microsoft's GraphRAG
\citep{edge2024graphrag} and LightRAG \citep{guo2024lightrag} use
shallow heuristic traversal over the resulting graph, which is fast but
loses much of the multi-hop signal. HippoRAG \citep{gutierrez2024hipporag}
and its successor HippoRAG-v2 \citep{gutierrez2025hipporag2} take the
stronger route of extracting named entities from the query, linking them
to nodes in a global corpus graph, and using those nodes to seed
Personalized PageRank (PPR)
\citep{page1999pagerank,haveliwala2002topicppr}. This provides effective
associative retrieval, but it introduces an LLM call before every graph
search and makes retrieval depend on the entities retained by that call.
Predicate-level cues that are not expressed as salient named entities may
therefore receive weak or no initial activation, causing relevant
relations to be overlooked. ActionRAG instead retrieves both entity
and concrete triple seeds directly, then uses local graph diffusion to
retroevaluate the resulting triples. This removes query-time LLM parsing
from retrieval while preserving relational cues that entity-only seeding
can miss.

Although an agent's memory may be large, evidence relevant to a
particular query often occupies a small region once plausible entities
and relations have been identified. ActionRAG builds this region over
a cautiously indexed entity--relation graph. Its indexing pipeline
reconciles type labels, resolves repeated or ambiguous entities through
tiered checks, and preserves source provenance. At query time, direct
retrieval over both entities and concrete triples produces an initial
evidence set. ActionRAG then
constructs a bounded local graph and applies query-conditioned,
receiver-normalized diffusion to revise triple relevance using
surrounding path structure. Graph connectivity therefore acts as a
second-stage evaluator of retrieved evidence rather than replacing
semantic retrieval altogether. We evaluate ActionRAG on the eight
deterministic-scoring columns of MemoryAgentBench. \textcolor{black}{ActionRAG
has the highest reported score among retrieval-based memory agents on six of
the eight columns where the unweighted mean of these heterogeneous scores is
$55.1$, $5.4$ points above HippoRAG-v2.}

\noindent\textbf{Contributions.}
\begin{itemize}
  \item A cautious graph-indexing pipeline with label reconciliation,
    tiered entity resolution, provenance preservation, and conflict-aware
    summarization.
  \item An evidence-refining local graph diffusion method that combines
    dual-channel retrieval with relational retroevaluation,
    \textcolor{black}{yielding the highest reported score among retrieval-based
    memory agents on six of the eight deterministic MAB columns.}
\end{itemize}

\section{Related Work}

\subsection{Long-Context and Memory-Augmented LLMs}

One line of work expands the context window to hold an entire interaction
history. Frontier models offer 128K--1M token contexts
\citep{openai2024gpt4o,anthropic2024claude35,google2024gemini15}, while
position interpolation \citep{chen2023positionintp}, NTK and YaRN scaling
\citep{peng2024yarn}, and LongRoPE \citep{ding2024longrope} extend them
further. However, accuracy declines when relevant information lies deep
in the window \citep{liu2024lostmiddle,kamradt2023niah}, and inference
cost rises with context length, so long context alone is insufficient.

External-memory systems instead store information around the LLM. MemGPT
\citep{packer2024memgptllmsoperatingsystems} pages between a small in-context working memory
and a larger archive, a design productionised by Letta \citep{letta2024}.
Mem0 and Mem0g \citep{chhikara2025mem0} extract structured user facts and
retrieve them through a learned policy. A-MEM \citep{xu2025amem} adds
agentic links and a self-organising memory graph and Zep
\citep{rasmussen2025zep} maintains a temporal knowledge graph for evolving
facts. Agentic retrieval systems such as Self-RAG \citep{asai2023selfrag}
and MIRIX \citep{wang2025mirix} add reflection or multi-agent memory
control around retrieval. These systems address the competencies defined
by MemoryAgentBench \citep{hu2025memoryagentbench}: accurate retrieval,
test-time learning, long-range understanding, and conflict resolution.
Their retrieval can nevertheless depend on opaque learned policies or
flat similarity lookups that underuse structural signal.

\subsection{Retrieval-Augmented Generation}

Retrieval-augmented generation \citep{lewis2020rag,guu2020realm}
conditions an LLM on top-$k$ documents retrieved by a sparse model such
as BM25 \citep{robertson2009bm25} or a dense bi-encoder
\citep{karpukhin2020dpr,izacard2022contriever,wang2022e5,xiao2024cpack}.
Recent embedding baselines include proprietary OpenAI Text-Embedding-3
models \citep{openai2024embeddings} and Qwen3-Embedding
\citep{zhang2025qwen3embedding}. Stronger pipelines add cross-encoder
or LLM re-rankers
\citep{nogueira2020monot5,sun2023rankgpt}, or late-interaction models
such as ColBERT and ColBERTv2
\citep{khattab2020colbert,santhanam2022colbertv2}. On MemoryAgentBench,
dense RAG is competitive for accurate retrieval but underperforms on
multi-hop and conflict-resolution questions. Independent chunk scoring
misses cross-chunk structure and cannot reliably distinguish
semantically similar statements when one supersedes the other.

\subsection{Knowledge Graphs for Retrieval and Reasoning}

A complementary line structures retrieval through graphs or persistent
memory. RAPTOR \citep{sarthi2024raptor} builds a tree of recursive
summaries, GraphRAG \citep{edge2024graphrag} extracts an entity--relation
graph and summarises its communities for global queries, and LightRAG
\citep{guo2024lightrag} combines local entity and global relation
retrieval over a lightweight graph. Cognee \citep{markovic2025cognee}
optimises the interface between knowledge graphs and LLMs for multi-hop
reasoning. These approaches can underrepresent specific relational paths
through summarization or heuristic expansion.
Mem0 \citep{chhikara2025mem0} extracts and consolidates salient
conversational memories, and its graph variant represents entities and
relations to support relational retrieval. A-MEM \citep{xu2025amem}
organises memories as Zettelkasten-inspired notes with generated
attributes and dynamic links, updating existing notes as new evidence
arrives. MemoRAG \citep{10.1145/3696410.3714805} instead uses a dual-system
architecture: a lightweight long-range model forms global memory and
generates draft-answer clues that guide retrieval for a more expressive
answering model.

HippoRAG \citep{gutierrez2024hipporag} seeds Personalized PageRank
\citep{page1999pagerank,haveliwala2002topicppr} over a corpus-level
entity graph using query-relevant entities, then aggregates propagated
scores to passages. HippoRAG-v2 \citep{gutierrez2025hipporag2} strengthens
passage integration and the online use of an LLM. More broadly, graph
diffusion propagates local relevance through weighted transition
powers, with Personalized PageRank and heat kernels as common instances
\citep{klicpera2019gdc}. Collectively, these methods span community
summarization, dynamic memory linking, memory-guided clue generation,
and global graph propagation, with trade-offs in query-time model
calls, graph-wide computation, and dependence on preselected entity
cues.

\section{ActionRAG}\label{sec:method}

\subsection{Overview}\label{sec:overview}

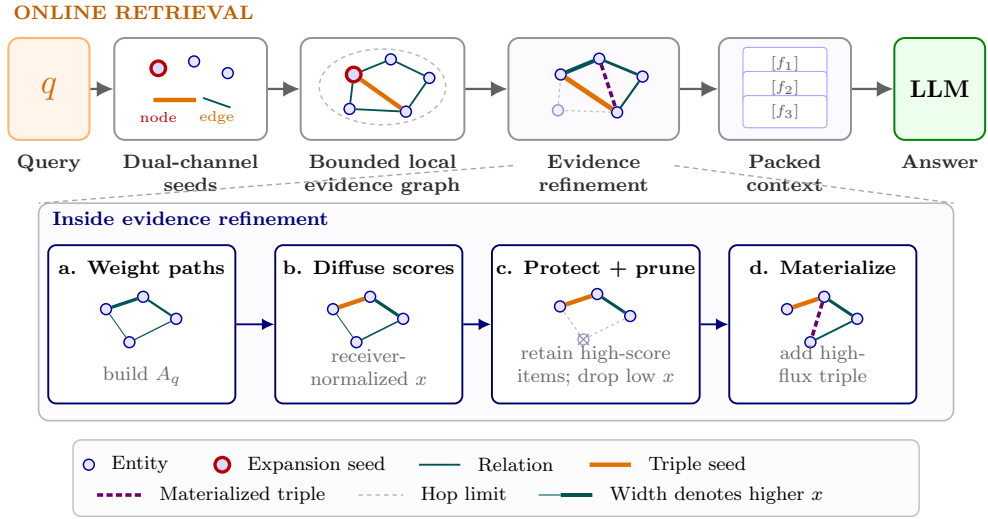
\begin{figure*}[t]
\centering
\begin{tikzpicture}[
    scale=0.91,
    every node/.style={font=\small, transform shape},
    card/.style={
      draw=black!42, line width=0.75pt, rounded corners=4pt,
      minimum height=1.48cm, minimum width=2.15cm,
      fill=white, inner sep=3pt
    },
    mainlabel/.style={
      font=\scriptsize\bfseries, align=center, text=black!82
    },
    entity/.style={
      circle, draw=blue!62!black, fill=blue!10,
      line width=0.55pt, minimum size=4.5pt, inner sep=0pt
    },
    seedentity/.style={
      circle, draw=red!72!black, fill=blue!14,
      line width=1.25pt, minimum size=6.0pt, inner sep=0pt
    },
    relation/.style={draw=teal!66!black, line width=0.70pt},
    relseed/.style={draw=orange!88!black, line width=1.65pt},
    materialedge/.style={
      draw=violet!82!black, line width=1.35pt,
      dash pattern=on 2.3pt off 1.25pt
    },
    lowedge/.style={
      draw=black!25, line width=0.45pt,
      dash pattern=on 1.1pt off 1.2pt
    },
    hopbnd/.style={
      draw=black!28, line width=0.55pt,
      dash pattern=on 1.7pt off 1.5pt
    },
    flow/.style={
      -{Latex[length=2.2mm,width=2.1mm]},
      draw=black!62, line width=0.85pt
    },
    substep/.style={
      draw=blue!42!black, line width=0.72pt, rounded corners=3pt,
      fill=white, minimum width=2.72cm, minimum height=2.30cm,
      inner sep=2pt
    },
    subarrow/.style={
      -{Latex[length=1.65mm,width=1.6mm]},
      draw=blue!48!black, line width=0.72pt
    },
]

\node[font=\scriptsize\bfseries, text=orange!75!black, anchor=west]
  at (-0.05,1.12) {ONLINE RETRIEVAL};

\node[card, minimum width=1.18cm, fill=orange!8, draw=orange!58]
  (R1) at (0.60,0) {};
\node[font=\Large\bfseries, text=orange!82!black] at (R1) {\(q\)};

\node[card, minimum width=2.22cm] (R2) at (2.66,0) {};
\begin{scope}[shift={($(R2)+(0,0.07)$)}]
  \node[seedentity] at (-0.47,0.23) {};
  \node[entity] at (0.05,0.33) {};
  \node[entity] at (0.54,0.16) {};
  \draw[relseed] (-0.54,-0.23) -- (0.08,-0.23);
  \draw[relation] (0.18,-0.20) -- (0.58,-0.34);
  \node[font=\tiny, text=red!68!black] at (-0.47,-0.48) {node};
  \node[font=\tiny, text=orange!82!black] at (0.38,-0.48) {edge};
\end{scope}

\node[card, minimum width=2.38cm] (R3) at (5.45,0) {};
\begin{scope}[shift={($(R3)+(0,0.03)$)}]
  \draw[hopbnd] (0,0) ellipse (0.92 and 0.55);
  \node[seedentity] (r3a) at (-0.42,0.18) {};
  \node[entity] (r3b) at ( 0.12,0.40) {};
  \node[entity] (r3c) at ( 0.68,0.13) {};
  \node[entity] (r3d) at ( 0.34,-0.36) {};
  \node[entity] (r3e) at (-0.48,-0.32) {};
  \draw[relation] (r3a) -- (r3b);
  \draw[relation] (r3b) -- (r3c);
  \draw[relseed] (r3a) -- (r3d);
  \draw[relation] (r3c) -- (r3d);
  \draw[relation] (r3d) -- (r3e);
  \draw[relation] (r3e) -- (r3a);
\end{scope}

\node[card, minimum width=2.48cm, fill=violet!2] (R4) at (8.52,0) {};
\begin{scope}[shift={($(R4)+(0,0.03)$)}]
  \node[entity] (r4a) at (-0.48,0.18) {};
  \node[entity] (r4b) at ( 0.08,0.41) {};
  \node[entity] (r4c) at ( 0.71,0.15) {};
  \node[entity] (r4d) at ( 0.34,-0.37) {};
  \node[entity, opacity=0.42] (r4e) at (-0.52,-0.34) {};
  \draw[relation, line width=1.55pt] (r4a) -- (r4b);
  \draw[relation, line width=1.05pt] (r4b) -- (r4c);
  \draw[relseed, line width=1.85pt] (r4a) -- (r4d);
  \draw[relation, line width=0.82pt] (r4c) -- (r4d);
  \draw[lowedge] (r4d) -- (r4e);
  \draw[lowedge] (r4e) -- (r4a);
  \draw[materialedge] (r4b) -- (r4d);
\end{scope}

\node[card, minimum width=1.92cm, fill=blue!2] (R5) at (11.30,0) {};
\begin{scope}[shift={($(R5)+(0,0.02)$)}]
  \node[draw=blue!35, fill=white, rounded corners=1.4pt,
        minimum width=1.22cm, minimum height=0.24cm,
        font=\tiny, text=black!65] at (0,0.35) {\([f_1]\)};
  \node[draw=blue!35, fill=white, rounded corners=1.4pt,
        minimum width=1.22cm, minimum height=0.24cm,
        font=\tiny, text=black!65] at (0,0.00) {\([f_2]\)};
  \node[draw=blue!35, fill=white, rounded corners=1.4pt,
        minimum width=1.22cm, minimum height=0.24cm,
        font=\tiny, text=black!65] at (0,-0.35) {\([f_3]\)};
\end{scope}

\node[card, minimum width=1.32cm, fill=green!7, draw=green!52!black]
  (R6) at (13.55,0) {\textbf{LLM}};

\draw[flow] (R1) -- (R2);
\draw[flow] (R2) -- (R3);
\draw[flow] (R3) -- (R4);
\draw[flow] (R4) -- (R5);
\draw[flow] (R5) -- (R6);

\node[mainlabel, below=2pt of R1] {Query};
\node[mainlabel, below=2pt of R2] {Dual-channel\\seeds};
\node[mainlabel, below=2pt of R3] {Bounded local\\evidence graph};
\node[mainlabel, below=2pt of R4] {Evidence\\refinement};
\node[mainlabel, below=2pt of R5] {Packed\\context};
\node[mainlabel, below=2pt of R6] {Answer};

\coordinate (insetTL) at (0.45,-1.66);
\coordinate (insetTR) at (13.75,-1.66);
\coordinate (insetBR) at (13.75,-4.82);
\coordinate (insetBL) at (0.45,-4.82);

\draw[black!27, rounded corners=4pt, fill=blue!1, line width=0.62pt]
  (insetTL) rectangle (insetBR);
\draw[black!35, dash pattern=on 2pt off 1.5pt, line width=0.52pt]
  ($(R4.south west)+(0.08,-0.36)$) -- (insetTL);
\draw[black!35, dash pattern=on 2pt off 1.5pt, line width=0.52pt]
  ($(R4.south east)+(-0.08,-0.36)$) -- (insetTR);

\node[anchor=north west, font=\scriptsize\bfseries, text=blue!48!black,
      fill=blue!1, inner sep=1.5pt]
  at ($(insetTL)+(0.14,-0.06)$) {Inside evidence refinement};

\node[substep] (S1) at (1.95,-3.42) {};
\node[substep] (S2) at (5.25,-3.42) {};
\node[substep, minimum width=3.02cm] (S3) at (8.55,-3.42) {};
\node[substep] (S4) at (11.85,-3.42) {};

\node[font=\scriptsize\bfseries, anchor=north] at ($(S1.north)+(0,-0.10)$)
  {a. Weight paths};
\node[font=\scriptsize\bfseries, anchor=north] at ($(S2.north)+(0,-0.10)$)
  {b. Diffuse scores};
\node[font=\scriptsize\bfseries, anchor=north] at ($(S3.north)+(0,-0.10)$)
  {\mbox{c. Protect + prune}};
\node[font=\scriptsize\bfseries, anchor=north] at ($(S4.north)+(0,-0.10)$)
  {d. Materialize};

\begin{scope}[shift={($(S1)+(0,0.06)$)}]
  \node[entity] (a1) at (-0.52,0.16) {};
  \node[entity] (a2) at ( 0.02,0.34) {};
  \node[entity] (a3) at ( 0.50,0.02) {};
  \node[entity] (a4) at (-0.18,-0.32) {};
  \draw[relation, line width=1.25pt] (a1) -- (a2);
  \draw[relation, line width=0.85pt] (a2) -- (a3);
  \draw[relation, line width=0.48pt] (a3) -- (a4);
  \draw[relation, line width=0.32pt] (a4) -- (a1);
\end{scope}
\node[font=\scriptsize, text=black!58, anchor=south,
      text width=2.42cm, align=center]
  at ($(S1.south)+(0,0.14)$) {build \(A_q\)};

\begin{scope}[shift={($(S2)+(0,0.06)$)}]
  \node[entity] (b1) at (-0.52,0.16) {};
  \node[entity] (b2) at ( 0.02,0.34) {};
  \node[entity] (b3) at ( 0.50,0.02) {};
  \node[entity] (b4) at (-0.18,-0.32) {};
  \draw[relseed, line width=1.65pt] (b1) -- (b2);
  \draw[relation, line width=1.20pt] (b2) -- (b3);
  \draw[relation, line width=0.60pt] (b3) -- (b4);
  \draw[relation, line width=0.32pt] (b4) -- (b1);
\end{scope}
\node[font=\scriptsize, text=black!58, anchor=south,
      text width=2.42cm, align=center]
  at ($(S2.south)+(0,0.14)$) {receiver-normalized \(x\)};

\begin{scope}[shift={($(S3)+(0,0.10)$)}]
  \node[entity] (c1) at (-0.52,0.16) {};
  \node[entity] (c2) at ( 0.02,0.34) {};
  \node[entity] (c3) at ( 0.50,0.02) {};
  \node[entity, opacity=0.38] (c4) at (-0.18,-0.32) {};
  \draw[relseed, line width=1.65pt] (c1) -- (c2);
  \draw[relation, line width=1.15pt] (c2) -- (c3);
  \draw[lowedge] (c3) -- (c4);
  \draw[lowedge] (c4) -- (c1);
  \draw[black!48, line width=0.35pt]
    ($(c4)+(-0.08,-0.08)$) -- ($(c4)+(0.08,0.08)$);
  \draw[black!48, line width=0.35pt]
    ($(c4)+(-0.08,0.08)$) -- ($(c4)+(0.08,-0.08)$);
\end{scope}
\node[font=\scriptsize, text=black!58, anchor=south,
      text width=2.42cm, align=center]
  at ($(S3.south)+(0,0.14)$) {retain high-score items; drop low \(x\)};

\begin{scope}[shift={($(S4)+(0,0.06)$)}]
  \node[entity] (d1) at (-0.52,0.16) {};
  \node[entity] (d2) at ( 0.02,0.34) {};
  \node[entity] (d3) at ( 0.50,0.02) {};
  \node[entity] (d4) at (-0.18,-0.32) {};
  \draw[relseed, line width=1.65pt] (d1) -- (d2);
  \draw[relation, line width=1.15pt] (d2) -- (d3);
  \draw[relation, line width=0.72pt] (d3) -- (d4);
  \draw[materialedge] (d2) -- (d4);
\end{scope}
\node[font=\scriptsize, text=black!58, anchor=south,
      text width=2.42cm, align=center]
  at ($(S4.south)+(0,0.14)$) {add high-flux triple};

\draw[subarrow] (S1) -- (S2);
\draw[subarrow] (S2) -- (S3);
\draw[subarrow] (S3) -- (S4);

\draw[black!24, rounded corners=3pt, fill=black!1, line width=0.55pt]
  (0.95,-5.10) rectangle (13.25,-6.21);

\node[entity] (le1) at (1.18,-5.46) {};
\node[right=3pt of le1, font=\scriptsize] {Entity};
\node[seedentity] (le2) at (3.12,-5.46) {};
\node[right=3pt of le2, font=\scriptsize] {Expansion seed};
\draw[relation] (5.98,-5.46) -- (6.58,-5.46);
\node[anchor=west, font=\scriptsize] at (6.70,-5.46) {Relation};
\draw[relseed] (8.46,-5.46) -- (9.06,-5.46);
\node[anchor=west, font=\scriptsize] at (9.18,-5.46) {Triple seed};

\draw[materialedge] (1.30,-5.90) -- (1.95,-5.90);
\node[anchor=west, font=\scriptsize] at (2.07,-5.90)
  {Materialized triple};
\draw[hopbnd] (5.10,-5.90) -- (5.75,-5.90);
\node[anchor=west, font=\scriptsize] at (5.87,-5.90) {Hop limit};
\draw[relation, line width=0.40pt] (7.72,-5.90) -- (8.04,-5.90);
\draw[relation, line width=1.50pt] (8.04,-5.90) -- (8.50,-5.90);
\node[anchor=west, font=\scriptsize] at (8.62,-5.90)
  {Width denotes higher \(x\)};

\end{tikzpicture}
\caption{ActionRAG online retrieval. A query retrieves entity nodes
as expansion anchors and concrete triple edges as evidence items, then
expands a bounded local evidence graph. Receiver-normalized diffusion
scores the item set of concrete triples plus orphan entities; high raw-score
items are protected, low-scoring evidence is pruned, and high-flux missing
triples are materialized before the surviving facts are packed into the
LLM context. Edge width denotes the diffused triple score.}
\label{fig:pipeline}
\end{figure*}

ActionRAG separates persistent memory construction from query-local
retrieval. It assumes that each query touches only a bounded
neighborhood of the accumulated interaction graph, so evidence
diffusion operates on a query-conditioned subgraph rather than the
corpus-level graph. Figure~\ref{fig:pipeline} summarizes the online
pipeline.

\textbf{Offline indexing} (\S\ref{sec:indexing}) converts multi-turn
interaction histories into a typed entity--relation graph. An LLM
extracts typed $(\text{head},\text{relation},\text{tail})$ triples,
entity mentions are canonicalized, and entity and triple descriptions
are embedded with BGE-M3 \citep{chen2024bgem3}. Postgres stores the
graph, its pgvector indices, and source-chunk back-pointers for every
triple.

\textbf{Online retrieval} (\S\ref{sec:online}) encodes the query once
and retrieves the top-$K_E$ entities and top-$K_T$ concrete triples in
parallel. Hop-capped expansion induces a bounded item graph, and
receiver-normalized diffusion refines its scores through the closed-form
solve $x=\alpha(I-(1-\alpha)P)^{-1}s$. Seed protection and pruning
preserve high raw-similarity evidence, while hub-penalized flux
materializes missing triples. The ordered evidence items are mapped to
their source chunks and packed under a token budget as
$[\text{fact}\ \#N]$ entries for answer generation.

\paragraph{Notation.}
We write the typed graph as $G = (E, R, T)$ with entity set $E$,
relation-type set $R$, and typed triple set
$T \subseteq E \times R \times E$. For a query $q$, the retrieval-time
item set $V \subseteq T \cup E$ contains the triples surfaced by seed
expansion together with any orphan entities that lack outgoing
relations in the local subgraph. Online ``relation'' retrieval therefore
means retrieval of concrete triples in $T$, not relation types in $R$.
We denote the nonnegative raw query-similarity scores by
$s \in \mathbb{R}_{\ge0}^{|V|}$ and the refined scores by
$x \in \mathbb{R}^{|V|}$. Let
$A \in \mathbb{R}^{|V| \times |V|}$ be the symmetric path-weighted
adjacency, $D = \mathrm{diag}(A\mathbf{1})$, and $P = D^{-1}A$ the
receiver-normalized propagation operator. The coefficient
$\alpha \in (0,1)$ retains raw seed evidence. The exact form of
$A_{ij}$, the hop cap, and the hub penalty are defined in
\S\ref{sec:online}.

\subsection{Cautious Graph Construction}\label{sec:indexing}

The indexer turns an interaction history into an entity--relation graph
and records the source chunk for every graph item. Chunks are processed
in parallel, but names and types are resolved across the complete batch
before the graph is updated.

\paragraph{Structured extraction and temporal grounding.}
For each chunk, an LLM emits typed entity mentions and directed
$(\text{head},\text{relation},\text{tail})$ mentions with textual
descriptions. The current type vocabulary is supplied as a soft
constraint, allowing the extractor to introduce a more specific label
when no existing type is suitable. Dates, durations, and temporal
phrases are retained inside the corresponding entity or relation
description rather than represented as graph nodes. When a dialogue
has a session timestamp, relative phrases such as ``yesterday'' or
``last week'' are converted to absolute dates using that timestamp.
The resolved date is stored in the fact description instead of creating
a separate node for it.

\paragraph{Label reconciliation.}
Free-form extraction improves coverage but can produce several names
for the same entity type. Before updating the graph, the indexer groups
proposed types using name and description similarity, maps each group
to an existing canonical type when possible, and registers unmatched
groups as new types. Alias chains are resolved before the resulting type
map is applied to the batch.

\paragraph{Entity resolution.}
Mentions are resolved after their types have been canonicalized. A
unique node with the same normalized name and a compatible type is
reused directly. If no exact match exists, the indexer reuses a fuzzy
name match only when it is the clear best candidate. All remaining
mentions are grouped by normalized name, canonical type, and description
similarity. One node is then created for each group before any relations
are written, preventing parallel workers from creating duplicate nodes.
Resolved entity pairs are stored as typed triples, with repeated
observations merged into the same edge. Each graph item retains links to
its source chunks, preserving the original evidence.

\paragraph{Temporal graph summarization.}
Nodes and relations supported by several mentions are periodically
summarized into a single description. Hierarchical summarization retains
dates and links to the source chunks, after which the updated description
is embedded for retrieval. The graph thus provides a compact semantic
representation while the original evidence remains available through
these links.

\subsection{Query-Time Evidence Refinement by Graph Diffusion}
\label{sec:online}

\paragraph{Motivation.}
Semantic retrieval scores graph items independently. It can therefore
retrieve two strongly matching facts while missing the relation that
connects them, simply because the connecting relation is not itself
similar to the query. Unrestricted graph expansion has the opposite
problem: high-degree entities introduce many generic facts and consume
the context budget. ActionRAG treats the initial retrieval as a
proposal rather than a final result. It constructs a bounded
query-local graph, retroevaluates each proposed item using nearby
evidence, removes weak proposals, and exposes useful relations that
were absent from the initial set.

\paragraph{Dual-channel proposal.}
Let $z=\phi(Q)$ be the embedding of query $Q$. Independent nearest
neighbor searches over entity and concrete triple descriptions recover
referential and predicate-level matches, respectively. Their union,
followed by bounded graph expansion, produces the initial item set
$\mathcal{V}_0=\{v_1,\ldots,v_n\}$, where an item is either a typed
triple or an orphan entity. Each item receives the nonnegative seed
score
\[
  s_i=\max(0,\cos(z,\phi(v_i))),
\]
giving the seed vector $s\in\mathbb{R}_{\ge0}^{n}$.

\paragraph{Query-conditioned item graph.}
A bounded entity graph $\mathcal{G}_Q$ is fetched around the endpoints
of $\mathcal{V}_0$. For affinity construction, we use the undirected
projection of the typed triple graph: a triple item has endpoint set
$\{h,t\}$, and an orphan entity item has endpoint set $\{e\}$. For each
concrete triple $\tau$ in this graph, let
\[
  r_\tau = \cos(z,\phi(\tau)),\qquad r_\tau^{+}=\max(0,r_\tau),
  \qquad
  c_\tau = \max(\delta,1-r_\tau),
\]
where $r_\tau$ measures query relevance and $c_\tau$ is its traversal cost.
For items $v_i$ and $v_j$, let $\pi_{ij}$ be the minimum-cost path
between any of their endpoints, subject to a hop bound $H$, and let
$d_{ij}=|\pi_{ij}|$. Its relevance is
$r_{ij}=\max_{\tau\in\pi_{ij}}r_\tau^{+}$, with $r_{ij}=1$ when the two
items share an endpoint. Shared endpoints are treated as direct local
evidence. High-degree noise is controlled by the expansion cutoff before
diffusion and by the hub penalty during materialization. These quantities
define the symmetric item affinity
\begin{equation}
  A_{ij} =
  \begin{cases}
    \exp(-\beta d_{ij})\,r_{ij},
      & i\neq j\ \text{and}\ d_{ij}\le H,\\
    0, & \text{otherwise}.
  \end{cases}
  \label{eq:item-affinity}
\end{equation}
Traversal costs select the path, while the final affinity uses the hop
length and the strongest positive triple relevance on that selected
path. Thus, two items interact strongly only when they are connected by
a short path with query-relevant evidence. This max-relevance choice lets
a salient bridge support diffusion even when nearby path triples mainly
serve as context carriers; traversal cost selects among candidate paths
and hop decay limits long bridges.

\paragraph{Receiver-normalized diffusion.}
Let $D=\operatorname{diag}(A\mathbf{1})$ and $P=D^{-1}A$, with an
all-zero row for an isolated item. ActionRAG refines the seed scores
through
\begin{equation}
  x=\alpha s+(1-\alpha)Px,
  \qquad
  x=\alpha\bigl(I-(1-\alpha)P\bigr)^{-1}s.
  \label{eq:receiver-diffusion}
\end{equation}
The orientation of $P$ is important. It acts directly on the column
vector $x$, so a non-isolated item receives
\begin{equation}
  x_i=\alpha s_i+
  (1-\alpha)\frac{\sum_j A_{ij}x_j}{\sum_k A_{ik}}.
  \label{eq:receiver-item}
\end{equation}
Each item therefore combines its own semantic evidence with a
relevance-weighted average of its neighbors. Normalization belongs to
the receiving item, which makes the operator an evidence smoother
rather than probability-mass transport. Since
$\rho((1-\alpha)P)<1$ for $\alpha\in(0,1)$, the fixed point is unique
and the closed-form solve is well defined.

\paragraph{Pruning and evidence materialization.}
Diffusion can weaken an isolated but highly query-relevant item, so
ActionRAG protects the strongest raw-similarity items. The retained
index set is
\begin{equation}
  m=\max_j x_j,\qquad
  \mathcal{K}=
  \begin{cases}
    \{i:x_i\ge\varepsilon m\}\cup\operatorname{Top}_{S}(s),
      & m>0,\\
    \operatorname{Top}_{S}(s), & m=0,
  \end{cases}
  \label{eq:seed-protection}
\end{equation}
where $S$ is the number of protected raw-score items. The second branch
is the all-zero fallback: if no item receives positive diffused score,
only the protected raw-score items are retained.
To recover triples that were not proposed initially, every concrete
triple $\tau\in\pi_{ij}$ between
retained items receives the flux score
\begin{equation}
  F_{ij\tau}=(x_i+x_j)\exp(-\beta d_{ij})r_\tau^{+}\,h_i h_j,
  \qquad
  h_i=\min\!\left(1,\frac{1}{\log\max(\Delta_i,2)}\right),
  \label{eq:flux}
\end{equation}
where $\Delta_i$ is the maximum local degree among the endpoints of
$v_i$. The factors $h_i$ and $h_j$ prevent hubs from dominating merely
because they participate in many paths. A unique unseen triple receives
the aggregate score
\[
  F_\tau=\max_{\substack{i,j\in\mathcal{K}\\ \tau\in\pi_{ij}}}F_{ij\tau}.
\]
The highest-flux triples absent from $\mathcal{V}_0$ are materialized as
additional evidence, up to a budget of $L$ triples.
This changes only the query-local candidate set, not the stored graph.

\begin{algorithm2e}[t]
  \small
  \DontPrintSemicolon
  \SetAlgoVlined
  \caption{Query-time evidence refinement and materialization.}
  \label{alg:diffuse}
  \KwIn{Query $Q$; graph $G$; embedder $\phi$; candidate parameters
    $K_E,K_T,D_E,D_T,K_{\text{init}}$; diffusion parameters
    $H,\alpha,\beta,\delta,\varepsilon,S,L$}
  \KwOut{Ordered query-local evidence set $\mathcal{V}^{*}$}
  $z\gets\phi(Q)$\;
  retrieve top-$K_E$ entity seeds and top-$K_T$ concrete triple seeds\;
  expand entity and triple seeds to depths $D_E$ and $D_T$, applying the
    fixed caps in Appendix~\ref{app:sentinel-config}, to obtain
    $(\mathcal{V}_0,s)$\;
  \lIf{$\mathcal{V}_0=\varnothing$}{\Return $[\,]$}
  \BlankLine
  fetch $\mathcal{G}_Q$ within $H$ hops of the item endpoints\;
  \ForEach{concrete triple $\tau\in\mathcal{G}_Q$}{
    compute its relevance $r_\tau$, positive relevance $r_\tau^{+}$, and
      traversal cost $c_\tau$\;
  }
  \ForEach{unordered item pair $(v_i,v_j)$}{
    find $\pi_{ij}$ and $d_{ij}$; set $A_{ij}$ using
      Eq.~\eqref{eq:item-affinity}\;
  }
  \BlankLine
  row-normalize $A$ to obtain $P$, using a zero row for each isolated
    item\;
  solve Eq.~\eqref{eq:receiver-diffusion} for $x$\;
  retain $\mathcal{K}$ according to Eq.~\eqref{eq:seed-protection}\;
  \BlankLine
  \ForEach{unseen triple $\tau$ on a path between retained items}{
    compute pairwise flux entries $F_{ij\tau}$ using Eq.~\eqref{eq:flux}\;
  }
  aggregate duplicate entries by $F_\tau=\max_{i,j}F_{ij\tau}$\;
  $\mathcal{M}\gets$ unique triples with the top-$L$ aggregate flux scores\;
  $\mathcal{V}^{*}\gets\{v_i:i\in\mathcal{K}\}\cup\mathcal{M}$\;
  \Return retained items ordered by $x_i$, followed by
    materialized triples ordered by $F_\tau$\;
\end{algorithm2e}

\paragraph{Context assembly.}
The evidence order is intentionally two stage: retained seed-supported
items are packed first by diffused score, then materialized bridge
triples are appended by flux. The retained and materialized items are
rendered as graph facts and greedily packed into the generator context
under a token budget. No LLM is invoked during retrieval, and all
materialized evidence is temporary to the current query.

\section{Experiments}\label{sec:experiments}

We evaluate ActionRAG on MemoryAgentBench
\citep{hu2025memoryagentbench}, which tests long-term memory through
incremental multi-turn interactions. We report its eight deterministic
task columns across four competencies: accurate retrieval (SH-DocQA,
MH-DocQA, and EventQA), conflict resolution (FC-SH and FC-MH),
long-range understanding (Detective QA), and test-time learning (MCC
and Recom). All \textcolor{black}{ActionRAG} results use the receiver-normalized
diffusion method in Section~\ref{sec:online}; unless stated otherwise, all
\textcolor{black}{baseline} numbers are taken directly from the MAB paper. We compare
against the MAB baseline families most relevant to retrieval-based
memory: long-context references, simple and embedding RAG,
structure-augmented RAG, and agentic memory agents, including
HippoRAG-v2 \citep{gutierrez2025hipporag2}.

\subsection{Setup}\label{sec:setup}

\paragraph{Benchmark and scoring.}
The eight columns aggregate finer-grained datasets: SH-DocQA and
MH-DocQA from RULER, three context-length variants of EventQA, five
in-context classification datasets for MCC, Redial for Recom,
Detective QA, and four context-length variants each for FC-SH and
FC-MH. Following MAB, we use substring exact match for DocQA, EventQA,
and fact consolidation, exact match for MCC and Detective QA, and
Recall@5 for Recom. We preserve MAB's within-column aggregation and
report the unweighted mean of the eight column scores as the
``deterministic average.'' Tasks requiring an LLM judge are excluded
from this aggregate.

\paragraph{ActionRAG configuration.}
Each query retrieves the ten nearest entities and ten nearest
concrete triples. Entity seeds are expanded by at most two hops and
triple seeds by one hop, after which the initial set is capped at
200 items.
The query-local graph uses hop bound $H{=}3$, distance-decay rate
$\beta{=}1$, and minimum traversal cost $\delta{=}0.01$. We apply one
receiver-normalized solve with direct-evidence weight $\alpha{=}0.15$;
no second diffusion round is performed. Items scoring below
$\varepsilon{=}0.10$ times the maximum refined score are pruned, except
for the $S{=}10$ strongest initial items, and at most $L{=}5$ unseen
triples are materialized using Eq.~\eqref{eq:flux}. Retrieved graph
facts are packed into a $7{,}000$-token context. Full retrieval settings
appear in Appendix~\ref{app:sentinel-config}.

Indexing follows Section~\ref{sec:indexing} and uses Claude Haiku~4.5
for extraction and graph summarization. BGE-M3 (${\sim}560$M
parameters) supplies all embeddings \citep{chen2024bgem3}, and
gpt-4o-mini generates the final answers for direct comparison with
MAB Table~17.

All ActionRAG experiments were run on a consumer Apple Silicon laptop
(macOS 26.5.2, Darwin 25.5.0, arm64). In comparison, MAB reports
HippoRAG-v2 timings on four NVIDIA L40 GPUs with an AMD EPYC 7713 CPU,
so the latency results are not hardware-matched.

\subsection{Main results on MAB deterministic columns}\label{sec:main-results}

\textcolor{black}{Table~\ref{tab:main} reports eight deterministic MAB columns
across long-context, RAG, structure-augmented, and agentic-memory baselines.
ActionRAG (three runs) is grouped as structure-augmented and uses MAB's
gpt-4o-mini generator.}

\begin{table}[t]
  \scriptsize\centering
  \setlength{\tabcolsep}{2.6pt}
  \caption{\textcolor{black}{Deterministic scores} on the selected MAB columns.
    Baseline numbers are from MAB Table~3. Unless a model is named in
    the agent row, MAB uses gpt-4o-mini as the answer generator. Avg.
    is the \textcolor{black}{unweighted mean of the eight heterogeneous
    displayed scores}. Bold marks
    the global column best and underline marks the best among
    structure-augmented RAG and agentic memory agents.}
  \label{tab:main}
  \begin{tabular}{@{}lrrrrrrrrr@{}}
    \toprule
                     & \multicolumn{3}{c}{AR} & \multicolumn{2}{c}{TTL}
                     & \multicolumn{1}{c}{LRU} & \multicolumn{2}{c}{CR}
                     & \multicolumn{1}{c}{Overall} \\
    \cmidrule(lr){2-4}\cmidrule(lr){5-6}\cmidrule(lr){7-7}
    \cmidrule(lr){8-9}\cmidrule(l){10-10}
    Agent type       & SH & MH & Event & MCC & Recom & DetQA & FC-SH & FC-MH & Avg. \\
    \midrule
    \multicolumn{10}{@{}l}{\emph{Long-context agents}}\\
    GPT-4o-mini (128K)    & 64.0 & 43.0 & 59.0 & \textbf{82.0} & 15.1 & \textbf{63.4} & 45.0 & 5.0 & 47.1 \\
    GPT-4.1-mini (1M)     & \textbf{83.0} & 66.0 & \textbf{82.6} & 75.6 & 16.7 & 56.3 & 36.0 & 5.0 & 52.7 \\
    \midrule
    \multicolumn{10}{@{}l}{\emph{Simple RAG agents}}\\
    BM25                 & 66.0 & 56.0 & 74.6 & 75.4 & 13.6 & 52.1 & 48.0 & 3.0 & 48.6 \\
    \midrule
    \multicolumn{10}{@{}l}{\emph{Embedding RAG agents}}\\
    Contriever           & 22.0 & 31.0 & 66.8 & 70.6 & 15.2 & 42.3 & 18.0 & 7.0 & 34.1 \\
    Text-Embed-3-Small   & 60.0 & 44.0 & 63.0 & 70.0 & 15.3 & 54.9 & 28.0 & 3.0 & 42.3 \\
    Text-Embed-3-Large   & 54.0 & 44.0 & 70.0 & 72.4 & 16.2 & 56.3 & 28.0 & 4.0 & 43.1 \\
    Qwen3-Embedding-4B   & 57.0 & 47.0 & 71.4 & 78.0 & 12.2 & 59.2 & 29.0 & 3.0 & 44.6 \\
    \midrule
    \multicolumn{10}{@{}l}{\emph{Structure-augmented RAG agents}}\\
    RAPTOR               & 29.0 & 38.0 & 45.8 & 59.4 & 12.3 & 42.3 & 14.0 & 1.0 & 30.2 \\
    GraphRAG             & 47.0 & 47.0 & 34.4 & 39.8 & 9.8 & 39.4 & 14.0 & 2.0 & 29.2 \\
    MemoRAG              & 29.0 & 33.0 & 56.0 & 77.0 & 13.1 & 50.7 & 21.0 & 7.0 & 35.9 \\
    HippoRAG-v2          & 76.0 & 66.0 & 67.6 & 61.4 & 10.2 & 57.7 & 54.0 & 5.0 & 49.7 \\
    Mem0                 & 25.0 & 32.0 & 37.5 & 32.4 & 10.0 & 36.6 & 18.0 & 2.0 & 24.2 \\
    Cognee               & 31.0 & 26.0 & 26.8 & 35.4 & 10.1 & 29.6 & 28.0 & 3.0 & 23.7 \\
    Zep                  & 44.0 & 25.0 & 42.5 & 62.8 & 12.1 & 28.2 & 7.0 & 3.0 & 28.1 \\
    \textcolor{black}{\shortstack[l]{\textbf{ActionRAG}\\[-1pt]\tiny\phantom{$\pm 1.0$}}} &
      \textcolor{black}{\shortstack{\underline{$77.0$}\\[-1pt]\tiny $\pm 1.0$}} &
      \textcolor{black}{\shortstack{$63.0$\\[-1pt]\tiny $\pm 1.0$}} &
      \textcolor{black}{\shortstack{\underline{$77.1$}\\[-1pt]\tiny $\pm 0.8$}} &
      \textcolor{black}{\shortstack{\underline{\textbf{$82.0$}}\\[-1pt]\tiny $\pm 1.1$}} &
      \textcolor{black}{\shortstack{\underline{\textbf{$18.2$}}\\[-1pt]\tiny $\pm 0.4$}} &
      \textcolor{black}{\shortstack{$46.7$\\[-1pt]\tiny $\pm 0.9$}} &
      \textcolor{black}{\shortstack{\underline{\textbf{$61.3$}}\\[-1pt]\tiny $\pm 1.2$}} &
      \textcolor{black}{\shortstack{\underline{\textbf{$15.5$}}\\[-1pt]\tiny $\pm 0.7$}} &
      \textcolor{black}{\shortstack{\underline{\textbf{$55.1$}}\\[-1pt]\tiny $\pm 0.9$}} \\
    \midrule
    \multicolumn{10}{@{}l}{\emph{Agentic memory agents}}\\
    Self-RAG             & 35.0 & 42.0 & 31.8 & 11.6 & 12.8 & 35.2 & 19.0 & 3.0 & 23.8 \\
    MemGPT               & 41.0 & 38.0 & 26.2 & 67.6 & 14.0 & 42.3 & 28.0 & 3.0 & 32.5 \\
    MIRIX                & 62.0 & 61.0 & 29.8 & 38.4 & 9.8 & 40.8 & 14.0 & 2.0 & 32.2 \\
    MIRIX (4.1-mini)     & 73.0 & \underline{\textbf{75.0}} & 53.0 & 61.0 & 10.3 & \underline{62.0} & 20.0 & 3.0 & 44.7 \\
    \bottomrule
  \end{tabular}
\end{table}

\begin{table}[t]
  \footnotesize\centering
  \begingroup\color{black}
  \setlength{\tabcolsep}{3.6pt}
  \caption{Component attribution on MAB deterministic aggregate.}
  \label{tab:component-attribution}
  \begin{tabular}{@{}lccrrr@{}}
    \toprule
    Graph construction & Recon. & Summ. & Base & PPR & ActionRAG \\
    \midrule
    Raw exact-match triplet graph & string only & none & 43.0 & 45.4 & 45.5 \\
    Summarized exact-match graph & string only & node/rel. & 43.8 & 46.0 & 48.9 \\
    Reconciled ActionRAG graph & node/type/rel. & node/rel.+prov. & 45.0 & 47.0 & 55.1 \\
    \bottomrule
  \end{tabular}
  \endgroup
\end{table}

\textcolor{black}{Table~\ref{tab:component-attribution} shows complementary
indexing and retrieval effects: the full pipeline scores $55.1$ on the
reconciled graph but $45.5$ on the raw graph and $48.9$ with summarization
without reconciliation. Base retrieval on the reconciled graph reaches
$45.0$. The best result therefore combines reconciled, summarized indexing
with query-local refinement.}

\textcolor{black}{ActionRAG has the highest reported retrieval-based score on
six columns and an unweighted mean of $55.1$, $5.4$ points above HippoRAG-v2.
MIRIX (4.1-mini) leads the combined structure-augmented/agentic group on
MH-DocQA and Detective QA, but uses iterative search and GPT-4.1-mini rather
than ActionRAG's single retrieval pass and gpt-4o-mini. Relative to
HippoRAG-v2, ActionRAG gains $20.6$, $8.0$, $7.3$, and $10.5$ points on MCC,
Recom, FC-SH, and FC-MH, but trails on MH-DocQA ($63.0$ vs. $66.0$) and
Detective QA ($46.7$ vs. $57.7$), where broader narrative context may be more
useful than a compact triple neighborhood.}

\textcolor{black}{Table~\ref{tab:ttl-detail} details MCC: ActionRAG nearly
matches GPT-4o-mini long context ($82.0$ vs. $82.4$), exceeds it on BANKING
and TREC-Coarse, and outperforms the reported retrieval and agentic-memory
baselines.}

\begin{table}[t]
  \scriptsize\centering
  \setlength{\tabcolsep}{2.4pt}
  \caption{\textcolor{black}{MCC test-time learning accuracy and the baselines are
    from MAB Table~7.}}
  \label{tab:ttl-detail}
  \begin{tabular}{@{}lrrrrrr@{}}
    \toprule
    Agent & BANKING & CLINIC & NLU & TREC-C & TREC-F & Avg. \\
    \midrule
    GPT-4o LC             & 96.0 & 96.0 & \textbf{90.0} & \textbf{87.0} & 69.0 & 87.6 \\
    Claude-3.7-Sonnet LC  & \textbf{97.0} & \textbf{98.0} & 86.0 & \textbf{87.0} & \textbf{79.0} & \textbf{89.4} \\
    GPT-4o-mini LC        & 93.0 & 93.0 & 87.0 & 73.0 & 66.0 & 82.4 \\
    Qwen3-Embedding-4B    & 90.0 & 88.0 & 86.0 & 67.0 & 59.0 & 78.0 \\
    MemoRAG               & 90.0 & 87.0 & 86.0 & 66.0 & 56.0 & 77.0 \\
    BM25                  & 89.0 & 89.0 & 84.0 & 62.0 & 53.0 & 75.4 \\
    MemGPT                & 89.0 & 83.0 & 79.0 & 56.0 & 31.0 & 67.6 \\
    Zep                   & 83.0 & 74.0 & 70.0 & 50.0 & 37.0 & 62.8 \\
    HippoRAG-v2           & 81.0 & 86.0 & 73.0 & 38.0 & 29.0 & 61.4 \\
    RAPTOR                & 78.0 & 75.0 & 73.0 & 48.0 & 23.0 & 59.4 \\
    GraphRAG              & 64.0 & 54.0 & 49.0 & 24.0 & 6.0  & 39.4 \\
    MIRIX (4o-mini)       & 42.0 & 53.0 & 49.0 & 36.0 & 12.0 & 38.4 \\
    Cognee                & 34.0 & 42.0 & 42.0 & 41.0 & 18.0 & 35.4 \\
    Mem0                  & 35.0 & 37.0 & 35.0 & 29.0 & 26.0 & 32.4 \\
    Self-RAG              & 19.0 & 13.0 & 6.0  & 15.0 & 5.0  & 11.6 \\
    \midrule
    \textbf{ActionRAG}  & 94.0 & 91.0 & 86.0 & 76.0 & 63.0 & 82.0 \\
    \bottomrule
  \end{tabular}
\end{table}

\noindent\textbf{Retrieval seed-count ablation.}
\textcolor{black}{Appendix~\ref{app:topk-detail} reports the exact MAB top-$k$
sweep on four columns. Here $k$ denotes retrieved chunks for MAB baselines and
$K_E{=}K_T{=}k$ entity and triple seeds for ActionRAG. The default $k=10$
matches Table~\ref{tab:main} and is ActionRAG's strongest tested setting.}

\subsection{Amortized per-query cost}\label{sec:cost}

\textcolor{black}{Following MAB Table~17, we use its November 2025 prices:
\$0.15/\$0.60 per 1M input/\allowbreak output tokens for GPT-4o-mini and
\$0.40/\$1.60 for GPT-4.1-mini. Baseline costs come from MAB; ActionRAG uses
measured mean tokens plus a 500-token prompt. This excludes graph
construction, storage, and local embedding.}

\begin{table}[t]
  \small\centering
  \setlength{\tabcolsep}{3.2pt}
  \caption{\textcolor{black}{Per-query cost (\$) / score on four MAB columns and the
    baselines are from MAB Table~17 under November 2025 prices.}}
  \label{tab:cost}
  \begin{tabular}{lcccc}
    \toprule
    Agent & MH-DocQA & MCC & DetQA & FC-SH \\
    \midrule
    GPT-4o-mini LC        & 0.010 / 43.0 & 0.008 / 82.0 & 0.010 / 63.4 & 0.010 / 45.0 \\
    GPT-4.1-mini LC       & 0.043 / 66.0 & 0.011 / 75.6 & 0.013 / 56.3 & 0.027 / 36.0 \\
    BM25 + gpt-4o-mini    & $<$0.001 / 56.0 & 0.006 / 75.4 & 0.006 / 52.1 & $<$0.001 / 48.0 \\
    MIRIX (4.1-mini)      & 0.016 / 75.0 & 0.010 / 61.0 & 0.011 / 62.0 & 0.019 / 20.0 \\
    \midrule
    \textbf{ActionRAG}  & 0.00109 / 63.0 & 0.00031 / 82.0 & 0.00116 / 46.7 & 0.00111 / 61.3 \\
    \bottomrule
  \end{tabular}
\end{table}

\FloatBarrier

\begin{samepage}
\textcolor{black}{ActionRAG costs at most \$0.00116/query. It matches
GPT-4o-mini long context on MCC at lower cost and on MH-DocQA/\allowbreak FC-SH it is
$39\times$/\allowbreak$17\times$ cheaper than GPT-4.1-mini/\allowbreak MIRIX, with a three-point
MH gap and higher FC-SH score.}
\end{samepage}

\noindent\textbf{Retrieval latency.}
\textcolor{black}{Excluding generation, retrieval takes $95.5$/$414.0$\,ms at
$p_{50}$/$p_{99}$ and BGE-M3 encoding $21.7$/$26.7$\,ms, a component-wise
$440.7$\,ms at $p_{99}$ (Appendix~\ref{app:latency-protocol}).}

\begin{samepage}
\section{Conclusion}\label{sec:conclusion}

\textcolor{black}{ActionRAG combines reconciled graph construction with
query-local receiver-normalized diffusion, pruning, and materialization for
long-term agent memory. Across eight heterogeneous deterministic MAB columns,
it records the highest reported retrieval-based score on six and an unweighted
mean of $55.1$. Component ablations indicate that the best configuration
combines reconciled, summarized indexing with retrieval-time refinement. Gains
center on test-time learning and conflict resolution where weaker MH-DocQA and
DetectiveQA results expose limits for narrative evidence. It also maintains
low query cost and sub-half-second main-evaluation retrieval, while packed
context remains nearly constant as the graph grows. Future work can include
better algorithms for graph expansion which are more robust to hyperparameter values.}
\end{samepage}


\bibliography{acml26}

\begin{thebibliography}{40}
\providecommand{\natexlab}[1]{#1}
\providecommand{\url}[1]{\texttt{#1}}
\expandafter\ifx\csname urlstyle\endcsname\relax
  \providecommand{\doi}[1]{doi: #1}\else
  \providecommand{\doi}{doi: \begingroup \urlstyle{rm}\Url}\fi

\bibitem[{Anthropic}(2024)]{anthropic2024claude35}
{Anthropic}.
\newblock {Claude} 3.5 {Sonnet}, 2024.
\newblock URL \url{https://www.anthropic.com/news/claude-3-5-sonnet}.

\bibitem[Asai et~al.(2024)Asai, Wu, Wang, Sil, and Hajishirzi]{asai2023selfrag}
Akari Asai, Zeqiu Wu, Yizhong Wang, Avirup Sil, and Hannaneh Hajishirzi.
\newblock {Self-RAG}: Learning to retrieve, generate, and critique through
  self-reflection.
\newblock In \emph{ICLR}, 2024.

\bibitem[Chen et~al.(2024)Chen, Xiao, Zhang, Luo, Lian, and Liu]{chen2024bgem3}
Jianlv Chen, Shitao Xiao, Peitian Zhang, Kun Luo, Defu Lian, and Zheng Liu.
\newblock {BGE M3-Embedding}: Multi-lingual, multi-functionality,
  multi-granularity text embeddings through self-knowledge distillation.
\newblock \emph{arXiv preprint arXiv:2402.03216}, 2024.

\bibitem[Chen et~al.(2023)Chen, Wong, Chen, and Tian]{chen2023positionintp}
Shouyuan Chen, Sherman Wong, Liangjian Chen, and Yuandong Tian.
\newblock Extending context window of large language models via positional
  interpolation.
\newblock \emph{arXiv preprint arXiv:2306.15595}, 2023.

\bibitem[Chhikara et~al.(2025)Chhikara, Khant, Aryan, Singh, and
  Yadav]{chhikara2025mem0}
Prateek Chhikara, Dev Khant, Saket Aryan, Taranjeet Singh, and Deshraj Yadav.
\newblock {Mem0}: Building production-ready {AI} agents with scalable long-term
  memory.
\newblock \emph{arXiv preprint arXiv:2504.19413}, 2025.

\bibitem[Ding et~al.(2024)Ding, Zhang, Zhang, Xu, Shang, Xu, Yang, and
  Yang]{ding2024longrope}
Yiran Ding, Li~Lyna Zhang, Chengruidong Zhang, Yuanyuan Xu, Ning Shang, Jiahang
  Xu, Fan Yang, and Mao Yang.
\newblock {LongRoPE}: Extending {LLM} context window beyond 2 million tokens.
\newblock In \emph{ICML}, 2024.

\bibitem[Edge et~al.(2024)Edge, Trinh, Cheng, Bradley, Chao, Mody, Truitt, and
  Larson]{edge2024graphrag}
Darren Edge, Ha~Trinh, Newman Cheng, Joshua Bradley, Alex Chao, Apurva Mody,
  Steven Truitt, and Jonathan Larson.
\newblock From local to global: A graph {RAG} approach to query-focused
  summarization.
\newblock \emph{arXiv preprint arXiv:2404.16130}, 2024.

\bibitem[{Gemini Team, Google}(2024)]{google2024gemini15}
{Gemini Team, Google}.
\newblock {Gemini 1.5}: Unlocking multimodal understanding across millions of
  tokens of context.
\newblock \emph{arXiv preprint arXiv:2403.05530}, 2024.

\bibitem[Guo et~al.(2024)Guo, Xia, Yu, Ao, and Huang]{guo2024lightrag}
Zirui Guo, Lianghao Xia, Yanhua Yu, Tu~Ao, and Chao Huang.
\newblock {LightRAG}: Simple and fast retrieval-augmented generation.
\newblock \emph{arXiv preprint arXiv:2410.05779}, 2024.

\bibitem[Guti{\'e}rrez et~al.(2024)Guti{\'e}rrez, Shu, Gu, Yasunaga, and
  Su]{gutierrez2024hipporag}
Bernal~Jim{\'e}nez Guti{\'e}rrez, Yiheng Shu, Yu~Gu, Michihiro Yasunaga, and
  Yu~Su.
\newblock {HippoRAG}: Neurobiologically inspired long-term memory for large
  language models.
\newblock In \emph{NeurIPS}, 2024.

\bibitem[Guti{\'e}rrez et~al.(2025)Guti{\'e}rrez, Shu, Qi, Zhou, and
  Su]{gutierrez2025hipporag2}
Bernal~Jim{\'e}nez Guti{\'e}rrez, Yiheng Shu, Weijian Qi, Sizhe Zhou, and
  Yu~Su.
\newblock From {RAG} to memory: Non-parametric continual learning for large
  language models.
\newblock \emph{arXiv preprint arXiv:2502.14802}, 2025.

\bibitem[Guu et~al.(2020)Guu, Lee, Tung, Pasupat, and Chang]{guu2020realm}
Kelvin Guu, Kenton Lee, Zora Tung, Panupong Pasupat, and Ming-Wei Chang.
\newblock {REALM}: Retrieval-augmented language model pre-training.
\newblock In \emph{ICML}, 2020.

\bibitem[Haveliwala(2002)]{haveliwala2002topicppr}
Taher~H. Haveliwala.
\newblock Topic-sensitive {PageRank}.
\newblock In \emph{WWW}, 2002.

\bibitem[Hu et~al.(2026)Hu, Wang, and McAuley]{hu2025memoryagentbench}
Yuanzhe Hu, Yu~Wang, and Julian McAuley.
\newblock Evaluating memory in {LLM} agents via incremental multi-turn
  interactions.
\newblock In \emph{ICLR}, 2026.

\bibitem[Izacard et~al.(2022)Izacard, Caron, Hosseini, Riedel, Bojanowski,
  Joulin, and Grave]{izacard2022contriever}
Gautier Izacard, Mathilde Caron, Lucas Hosseini, Sebastian Riedel, Piotr
  Bojanowski, Armand Joulin, and Edouard Grave.
\newblock Unsupervised dense information retrieval with contrastive learning.
\newblock \emph{Transactions on Machine Learning Research}, 2022.

\bibitem[Kamradt(2023)]{kamradt2023niah}
Greg Kamradt.
\newblock {Needle In A Haystack}: Pressure testing {LLM}s, 2023.
\newblock URL \url{https://github.com/gkamradt/LLMTest_NeedleInAHaystack}.

\bibitem[Karpukhin et~al.(2020)Karpukhin, O{\u{g}}uz, Min, Lewis, Wu, Edunov,
  Chen, and Yih]{karpukhin2020dpr}
Vladimir Karpukhin, Barlas O{\u{g}}uz, Sewon Min, Patrick Lewis, Ledell Wu,
  Sergey Edunov, Danqi Chen, and Wen-tau Yih.
\newblock Dense passage retrieval for open-domain question answering.
\newblock In \emph{EMNLP}, 2020.

\bibitem[Khattab and Zaharia(2020)]{khattab2020colbert}
Omar Khattab and Matei Zaharia.
\newblock {ColBERT}: Efficient and effective passage search via contextualized
  late interaction over {BERT}.
\newblock In \emph{SIGIR}, 2020.

\bibitem[Klicpera et~al.(2019)Klicpera, Wei{\ss}enberger, and
  G{\"u}nnemann]{klicpera2019gdc}
Johannes Klicpera, Stefan Wei{\ss}enberger, and Stephan G{\"u}nnemann.
\newblock Diffusion improves graph learning.
\newblock In \emph{NeurIPS}, 2019.

\bibitem[{Letta}(2024)]{letta2024}
{Letta}.
\newblock {Letta}: Stateful agents framework with memory and reasoning, 2024.
\newblock URL \url{https://www.letta.com/}.

\bibitem[Lewis et~al.(2020)Lewis, Perez, Piktus, Petroni, Karpukhin, Goyal,
  K{\"u}ttler, Lewis, Yih, Rockt{\"a}schel, Riedel, and Kiela]{lewis2020rag}
Patrick Lewis, Ethan Perez, Aleksandra Piktus, Fabio Petroni, Vladimir
  Karpukhin, Naman Goyal, Heinrich K{\"u}ttler, Mike Lewis, Wen-tau Yih, Tim
  Rockt{\"a}schel, Sebastian Riedel, and Douwe Kiela.
\newblock Retrieval-augmented generation for knowledge-intensive {NLP} tasks.
\newblock In \emph{NeurIPS}, 2020.

\bibitem[Liu et~al.(2024)Liu, Lin, Hewitt, Paranjape, Bevilacqua, Petroni, and
  Liang]{liu2024lostmiddle}
Nelson~F. Liu, Kevin Lin, John Hewitt, Ashwin Paranjape, Michele Bevilacqua,
  Fabio Petroni, and Percy Liang.
\newblock Lost in the middle: How language models use long contexts.
\newblock \emph{Transactions of the Association for Computational Linguistics},
  2024.

\bibitem[Markovic et~al.(2025)Markovic, Obradovic, Hajdu, and
  Pavlovic]{markovic2025cognee}
Vasilije Markovic, Lazar Obradovic, Laszlo Hajdu, and Jovan Pavlovic.
\newblock Optimizing the interface between knowledge graphs and {LLM}s for
  complex reasoning.
\newblock \emph{arXiv preprint arXiv:2505.24478}, 2025.

\bibitem[Nogueira et~al.(2020)Nogueira, Jiang, Pradeep, and
  Lin]{nogueira2020monot5}
Rodrigo Nogueira, Zhiying Jiang, Ronak Pradeep, and Jimmy Lin.
\newblock Document ranking with a pretrained sequence-to-sequence model.
\newblock \emph{Findings of EMNLP}, 2020.

\bibitem[{OpenAI}(2024{\natexlab{a}})]{openai2024embeddings}
{OpenAI}.
\newblock New embedding models and {API} updates, 2024{\natexlab{a}}.
\newblock URL
  \url{https://openai.com/index/new-embedding-models-and-api-updates/}.

\bibitem[{OpenAI}(2024{\natexlab{b}})]{openai2024gpt4o}
{OpenAI}.
\newblock {GPT-4o} system card, 2024{\natexlab{b}}.
\newblock URL \url{https://openai.com/index/gpt-4o-system-card/}.

\bibitem[Packer et~al.(2024)Packer, Wooders, Lin, Fang, Patil, Stoica, and
  Gonzalez]{packer2024memgptllmsoperatingsystems}
Charles Packer, Sarah Wooders, Kevin Lin, Vivian Fang, Shishir~G. Patil, Ion
  Stoica, and Joseph~E. Gonzalez.
\newblock {MemGPT}: Towards {LLM}s as operating systems, 2024.
\newblock URL \url{https://arxiv.org/abs/2310.08560}.

\bibitem[Page et~al.(1999)Page, Brin, Motwani, and Winograd]{page1999pagerank}
Lawrence Page, Sergey Brin, Rajeev Motwani, and Terry Winograd.
\newblock The {PageRank} citation ranking: Bringing order to the web.
\newblock Technical report, Stanford InfoLab, 1999.

\bibitem[Peng et~al.(2024)Peng, Quesnelle, Fan, and Shippole]{peng2024yarn}
Bowen Peng, Jeffrey Quesnelle, Honglu Fan, and Enrico Shippole.
\newblock {YaRN}: Efficient context window extension of large language models.
\newblock In \emph{ICLR}, 2024.

\bibitem[Qian et~al.(2025)Qian, Liu, Zhang, Mao, Lian, Dou, and
  Huang]{10.1145/3696410.3714805}
Hongjin Qian, Zheng Liu, Peitian Zhang, Kelong Mao, Defu Lian, Zhicheng Dou,
  and Tiejun Huang.
\newblock {MemoRAG}: Boosting long context processing with global
  memory-enhanced retrieval augmentation.
\newblock In \emph{Proceedings of the ACM on Web Conference 2025}, {WWW} '25,
  pages 2366--2377, New York, NY, USA, 2025. Association for Computing
  Machinery.
\newblock ISBN 9798400712746.
\newblock \doi{10.1145/3696410.3714805}.
\newblock URL \url{https://doi.org/10.1145/3696410.3714805}.

\bibitem[Rasmussen et~al.(2025)Rasmussen, Paliychuk, Beauvais, Ryan, and
  Chalef]{rasmussen2025zep}
Preston Rasmussen, Pavlo Paliychuk, Travis Beauvais, Jack Ryan, and Daniel
  Chalef.
\newblock {Zep}: A temporal knowledge graph architecture for agent memory.
\newblock \emph{arXiv preprint arXiv:2501.13956}, 2025.

\bibitem[Robertson and Zaragoza(2009)]{robertson2009bm25}
Stephen Robertson and Hugo Zaragoza.
\newblock The probabilistic relevance framework: {BM25} and beyond.
\newblock \emph{Foundations and Trends in Information Retrieval}, 3\penalty0
  (4):\penalty0 333--389, 2009.

\bibitem[Santhanam et~al.(2022)Santhanam, Khattab, Saad-Falcon, Potts, and
  Zaharia]{santhanam2022colbertv2}
Keshav Santhanam, Omar Khattab, Jon Saad-Falcon, Christopher Potts, and Matei
  Zaharia.
\newblock {ColBERTv2}: Effective and efficient retrieval via lightweight late
  interaction.
\newblock In \emph{NAACL}, 2022.

\bibitem[Sarthi et~al.(2024)Sarthi, Abdullah, Tuli, Khanna, Goldie, and
  Manning]{sarthi2024raptor}
Parth Sarthi, Salman Abdullah, Aditi Tuli, Shubh Khanna, Anna Goldie, and
  Christopher~D. Manning.
\newblock {RAPTOR}: Recursive abstractive processing for tree-organized
  retrieval.
\newblock In \emph{ICLR}, 2024.

\bibitem[Sun et~al.(2023)Sun, Yan, Ma, Wang, Ren, Chen, Yin, and
  Ren]{sun2023rankgpt}
Weiwei Sun, Lingyong Yan, Xinyu Ma, Shuaiqiang Wang, Pengjie Ren, Zhumin Chen,
  Dawei Yin, and Zhaochun Ren.
\newblock Is {ChatGPT} good at search? {Investigating} large language models as
  re-ranking agents.
\newblock \emph{EMNLP}, 2023.

\bibitem[Wang et~al.(2022)Wang, Yang, Huang, Jiao, Yang, Jiang, Majumder, and
  Wei]{wang2022e5}
Liang Wang, Nan Yang, Xiaolong Huang, Binxing Jiao, Linjun Yang, Daxin Jiang,
  Rangan Majumder, and Furu Wei.
\newblock Text embeddings by weakly-supervised contrastive pre-training.
\newblock \emph{arXiv preprint arXiv:2212.03533}, 2022.

\bibitem[Wang and Chen(2025)]{wang2025mirix}
Yu~Wang and Xi~Chen.
\newblock {MIRIX}: Multi-agent memory system for {LLM}-based agents.
\newblock \emph{arXiv preprint arXiv:2507.07957}, 2025.

\bibitem[Xiao et~al.(2024)Xiao, Liu, Zhang, Muennighoff, Lian, and
  Nie]{xiao2024cpack}
Shitao Xiao, Zheng Liu, Peitian Zhang, Niklas Muennighoff, Defu Lian, and
  Jian-Yun Nie.
\newblock {C-Pack}: Packed resources for general {C}hinese embeddings.
\newblock \emph{SIGIR}, 2024.

\bibitem[Xu et~al.(2025)Xu, Liang, Mei, Gao, Tan, and Zhang]{xu2025amem}
Wujiang Xu, Zujie Liang, Kai Mei, Hang Gao, Juntao Tan, and Yongfeng Zhang.
\newblock {A-MEM}: Agentic memory for {LLM} agents.
\newblock \emph{arXiv preprint arXiv:2502.12110}, 2025.

\bibitem[Zhang et~al.(2025)Zhang, Li, Long, Zhang, Lin, Yang, Xie, Yang, Liu,
  Lin, Huang, and Zhou]{zhang2025qwen3embedding}
Yanzhao Zhang, Mingxin Li, Dingkun Long, Xin Zhang, Huan Lin, Baosong Yang,
  Pengjun Xie, An~Yang, Dayiheng Liu, Junyang Lin, Fei Huang, and Jingren Zhou.
\newblock {Qwen3} embedding: Advancing text embedding and reranking through
  foundation models.
\newblock \emph{arXiv preprint arXiv:2506.05176}, 2025.

\end{thebibliography}

\appendix

\section{Retrieval top-$k$ ablation detail}\label{app:topk-detail}

\textcolor{black}{Table~\ref{tab:topk-detail} reports the top-$k$ sweep.
Baseline values are from MAB Table~9.}

\begin{table}[!htbp]
  \scriptsize\centering
  \setlength{\tabcolsep}{3.2pt}
  \caption{\textcolor{black}{Top-$k$ accuracy: $k$ is the chunk count for MAB
    baselines and $K_E{=}K_T{=}k$ for ActionRAG seeds.}}
  \label{tab:topk-detail}
  \begin{tabular}{llrrrr}
    \toprule
    Dataset & Method & $k{=}2$ & $k{=}5$ & $k{=}10$ \\
    \midrule
    SH-DocQA & BM25 & 50.0 & 60.0 & 66.0 \\
    SH-DocQA & Text-Embed-3-Large & 36.0 & 47.0 & 54.0 \\
    SH-DocQA & HippoRAG-v2 & 60.0 & 69.0 & 76.0 \\
    SH-DocQA & ActionRAG & 76.0 & 76.0 & 77.0 \\
    \midrule
    MH-DocQA & BM25 & 49.0 & 54.0 & 56.0 \\
    MH-DocQA & Text-Embed-3-Large & 37.0 & 41.0 & 44.0 \\
    MH-DocQA & HippoRAG-v2 & 53.0 & 60.0 & 66.0 \\
    MH-DocQA & ActionRAG & 56.0 & 57.0 & 63.0 \\
    \midrule
    EventQA & BM25 & 66.6 & 71.2 & 74.6 \\
    EventQA & Text-Embed-3-Large & 51.8 & 62.4 & 70.0 \\
    EventQA & HippoRAG-v2 & 58.8 & 67.6 & 67.4 \\
    EventQA & ActionRAG & 60.3 & 60.7 & 77.1 \\
    \midrule
    MCC & BM25 & 67.8 & 74.6 & 75.4 \\
    MCC & Text-Embed-3-Large & 59.4 & 69.4 & 72.4 \\
    MCC & HippoRAG-v2 & 58.8 & 61.4 & 61.4 \\
    MCC & ActionRAG & 81.6 & 80.4 & 82.0 \\
    \bottomrule
  \end{tabular}
\end{table}

\FloatBarrier

\section{\textcolor{black}{Additional Ablations and Sensitivity Analyses}}
\label{app:additional-rebuttal-experiments}

\begin{table}[!htbp]
  \scriptsize\centering
  \begingroup\color{black}
  \setlength{\tabcolsep}{3.1pt}
  \caption{One-at-a-time MCC sensitivity. Each row changes one default.}
  \label{tab:diffusion-sensitivity}
  \begin{tabular}{@{}lrrrrrr@{}}
    \toprule
    Config & BANK & CLINIC & NLU & TREC-C & TREC-F & MCC avg \\
    \midrule
    default ($\alpha=.15$, $H=3$, $\varepsilon=.1$, $L=5$, $S=10$)
      & 94.0 & 91.0 & 86.0 & 76.0 & 63.0 & 82.0 \\
    $\alpha=0.05$ & 94.0 & 90.0 & 85.0 & 82.0 & 63.0 & 82.8 \\
    $\alpha=0.50$ & 92.0 & 89.0 & 86.0 & 79.0 & 64.0 & 82.0 \\
    $\alpha=0.85$ & 95.0 & 89.0 & 86.0 & 86.0 & 68.0 & 84.8 \\
    $H=1$ & 93.0 & 88.0 & 87.0 & 83.0 & 63.0 & 82.8 \\
    $H=5$ & 92.0 & 89.0 & 87.0 & 80.0 & 66.0 & 82.8 \\
    $\varepsilon=0.0$ & 93.0 & 90.0 & 85.0 & 80.0 & 68.0 & 83.2 \\
    $\varepsilon=0.3$ & 93.0 & 89.0 & 85.0 & 84.0 & 62.0 & 82.6 \\
    $L=0$ & 93.0 & 88.0 & 87.0 & 84.0 & 65.0 & 83.4 \\
    $L=10$ & 92.0 & 89.0 & 87.0 & 80.0 & 63.0 & 82.2 \\
    $S=0$ & 93.0 & 89.0 & 87.0 & 80.0 & 62.0 & 82.2 \\
    $S=20$ & 93.0 & 89.0 & 86.0 & 80.0 & 63.0 & 82.2 \\
    \bottomrule
  \end{tabular}
  \endgroup
\end{table}

\textcolor{black}{MCC means span $82.0$--$84.8$. Both $\alpha=0.85$ ($84.8$)
and $L=0$ ($83.4$) exceed the fixed default, so it is not MCC-optimal and
materialization is not uniformly beneficial and all settings remain within
$2.8$ points.}

\begin{table}[!htbp]
  \scriptsize\centering
  \begingroup\color{black}
  \setlength{\tabcolsep}{4pt}
  \caption{MCC embedding sensitivity with fixed graph and retrieval settings.}
  \label{tab:embedding-sensitivity}
  \begin{tabular}{@{}lrrrrrrr@{}}
    \toprule
    Embedder & Dim & BANK & CLINIC & NLU & TREC-C & TREC-F & MCC avg \\
    \midrule
    BGE-M3 & 1024 & 94.0 & 89.0 & 86.0 & 84.0 & 67.0 & 84.0 \\
    text-embedding-3-large & 3072 & 92.0 & 91.0 & 88.0 & 74.0 & 60.0 & 81.0 \\
    text-embedding-3-small & 1536 & 90.0 & 92.0 & 87.0 & 78.0 & 51.0 & 79.6 \\
    \bottomrule
  \end{tabular}
  \endgroup
\end{table}

\FloatBarrier

\textcolor{black}{Each backend uses a fresh MCC run after rebuilding
embeddings, so BGE-M3 is separate from the main-table run. Its $84.0$ mean
leads text-embedding-3-large ($81.0$) and -3-small ($79.6$) with TREC-F varying
the most.}

\begin{table}[h]
  \tiny\centering
  \begingroup\color{black}
  \setlength{\tabcolsep}{2.5pt}
  \caption{EventQA graph growth on matched 64K, 128K, and full questions.}
  \label{tab:graph-expansion}
  \resizebox{\textwidth}{!}{%
  \begin{tabular}{@{}lrrrrrrrrrrrr@{}}
    \toprule
    Graph & Nodes & Tokens & Doc 1 & Doc 2 & Doc 3 & Doc 4 & Doc 5
    & Avg & $p_{50}$ ms & $p_{99}$ ms & Ctx tok & Items retr. \\
    \midrule
    64K & 409 & $\sim$77.9K & 72 & 77 & 73 & 74 & 70
      & 73.2 & 2212 & 5811 & 6844 & 103 \\
    128K & 775 & $\sim$156.8K & 69 & 75 & 69 & 69 & 70
      & 70.4 & 2233 & 6325 & 6901 & 121 \\
    full & 2948 & $\sim$651K & 75 & 76 & 63 & 67 & 64
      & 69.0 & 2764 & 7664 & 6896 & 153 \\
    $\Delta$ 64K to full & 7.2x & 8.4x & +3 & -1 & -10 & -7 & -6
      & -4.2 & +25\% & +32\% & $\sim$flat & +1.5x \\
    \bottomrule
  \end{tabular}}
  \endgroup
\end{table}

\textcolor{black}{From 64K to full EventQA, nodes grow $409\!\to\!2{,}948$ and
tokens $8.4\times$, while packed context stays near $6.9$K. Latency rises
$25\%$/$32\%$ at $p_{50}$/$p_{99}$ and mean score falls $4.2$ points as
distractors increase.}

\FloatBarrier

\section{ActionRAG configuration}\label{app:sentinel-config}

\textcolor{black}{All MAB columns use one fixed configuration: Claude
Haiku~4.5 extraction/summarization, BGE-M3 embeddings, and gpt-4o-mini
synthesis. Table~\ref{tab:config} maps implementation settings to
Section~\ref{sec:online}; implementation $K_R$ corresponds to paper $K_T$.}

\begin{table}[h]
  \scriptsize\centering
  \begingroup\color{black}
  \setlength{\tabcolsep}{3.2pt}
  \caption{Fixed ActionRAG retrieval configuration.}
  \label{tab:config}
  \begin{tabular}{@{}lll@{\qquad}lll@{}}
    \toprule
    Setting & Symbol & Value & Setting & Symbol & Value \\
    \midrule
    entity seeds & $K_E$ & 10 & hop cap & $H$ & 3 \\
    concrete triple seeds & $K_T$ & 10 & distance decay & $\beta$ & 1.0 \\
    entity expansion depth & $D_E$ & 2 & traversal floor & $\delta$ & 0.01 \\
    triple expansion depth & $D_T$ & 1 & direct-evidence weight & $\alpha$ & 0.15 \\
    neighbors per expansion & n/a & 8 & prune threshold & $\varepsilon$ & 0.10 \\
    hub-degree cutoff & n/a & 80 & protected raw-score items & $S$ & 10 \\
    initial item-pool cap & $K_{\text{init}}$ & 200 & materialized triples & $L$ & 5 \\
    final triple cap & n/a & 200 & retrieved-context budget & n/a & 7{,}000 tokens \\
    \bottomrule
  \end{tabular}
  \endgroup
\end{table}

\textcolor{black}{The implementation uses
$c_\tau=\max(0.01,1-\cos(z,\phi(\tau)))$ and one diffusion solve. In
\texttt{per\_intermediary} mode, unseen path triples receive pairwise flux and
duplicates are merged by maximum $F_\tau$, and the top $L$ become temporary
query-local evidence.}

\section{Retrieval latency protocol}\label{app:latency-protocol}

\textcolor{black}{We measure isolated retrieval, excluding answer generation,
on the same Apple Silicon laptop with local Postgres/pgvector and no discrete
GPU. Diffusion timing covers 50 MH-DocQA queries after initialization and
includes local-graph construction, path scoring, solving, pruning,
materialization, and packing, but excludes query embedding. BGE-M3 embedding
is measured separately at batch size one with normalized embeddings, 20
warm-ups, and 200 measured calls. The reported $p_{99}$ adds the two component
$p_{99}$ values and we omit concurrent full-retrieval wall time because database
and Python scheduling make it more variable.}

\end{document}